\documentclass{article} % For LaTeX2e
\usepackage{nips15submit_e,times}
\usepackage{hyperref}
\usepackage{url}
\usepackage{amsmath}
\usepackage{subcaption}
\usepackage{graphicx}
\usepackage{amsfonts}
\usepackage{booktabs}
\usepackage{stfloats}
\usepackage{float}
\usepackage{bbm}
\usepackage{placeins}

\usepackage{needspace}
\usepackage[font=small,skip=2pt]{caption}
\newcommand{\lossplot}[1]{\includegraphics[width=\linewidth,height=0.16\textheight,keepaspectratio]{#1}}
\newcommand{\barplot}[1]{\includegraphics[width=\linewidth,height=0.22\textheight,keepaspectratio]{#1}}

\title{In-Context Learning in Linear vs. Quadratic Attention Models: An Empirical Study on Regression Tasks}

\author{
Ayush Goel \\
University of California, Berkeley\\
\texttt{ayushgoel@berkeley.edu} \\
\And
Arjun Kohli \\
University of California, Berkeley \\
\texttt{kohli10@berkeley.edu} \\
\AND
Sarvagya Somvanshi \\
University of California, Berkeley \\
\texttt{sarvagya@berkeley.edu} \\
}

\nipsfinalcopy % Uncomment for camera-ready version

\begin{document}

\maketitle

\section*{Abstract}
Recent work has demonstrated that transformers and linear attention models can perform in-context learning (ICL) on simple function classes, such as linear regression. In this paper, we empirically study how these two attention mechanisms differ in their ICL behavior on the canonical linear-regression task of Garg et al. [3]. We evaluate learning quality (MSE), convergence, and generalization behavior of each architecture. We also analyze how increasing model depth affects ICL performance. Our results illustrate both the similarities and limitations of linear attention relative to quadratic attention in this setting.

\section{Introduction and Background}

In-context learning (ICL) emerged as a transformative capability when GPT-3 demonstrated that language models could act as few-shot learners without task-specific fine-tuning. This represented a fundamental shift from training individual models per task toward systems that exhibit meta-learning and can adapt to new functions at inference time.

The foundational work demonstrating ICL in controlled settings came from Garg et al. who were the first to show that transformers can be trained from scratch to perform in-context learning of entire function classes. Using linear regression as their core example, they showed that even a small transformer can learn unseen linear functions from in-context examples with accuracy very close to the optimal least squares solution. The transformer was also robust to distribution shifts, and exhibited similar traits when trained to learn more complex function classes like decision trees and small neural networks. This work established the linear regression ICL task as a canonical testbed for understanding transformer capabilities, which is the same benchmark we adopt in our comparative study of attention mechanisms.

Understanding \textit{why} transformers exhibit ICL required mechanistic analysis of their internal computations. Von Oswald et al. offered one of the first mechanistic explanations by demonstrating that a single layer of linear self-attention can implement the same data transformation as one step of gradient descent [8]. Building on this, they showed that the transformer as a whole trained on simple regression learns behaviors that mirror gradient descent performed on the forward pass when given a new prompt. This revealed that ICL emerges from gradient-based meta-learning rather than some fundamentally different mechanism.

Extending this mechanistic perspective, Akyürek et al. showed that transformers trained on linear regression tasks often learn internal mechanisms whose behavior corresponds to classical learning algorithms -- the resulting predictor can look similar to gradient descent, ridge regression or least squares regression [1]. Under different conditions, the transformer approximates different classical estimators rather than explicitly selecting among them. This work suggests that the choice of architecture and training conditions may determine which classical algorithm emerges, motivating our comparison of how quadratic versus linear attention might learn different algorithmic strategies.

While these mechanistic insights focused on standard softmax attention, alternative attention mechanisms offer different computational trade-offs and inductive biases. Katharopoulos et al. proposed linear attention as an alternative to standard quadratic self-attention by replacing the softmax kernel with a feature map that enables the attention computation to be expressed as a linear dot product [4]. They showed that this formulation gave rise to a recurrent-style computation, enabling transformers to process sequences of arbitrary length without revisiting prior inputs. Importantly for our setting, the choice of feature map directly determines the expressivity of the resulting attention mechanism: different kernels induce different inductive biases and therefore support different classes of input-output transformations. However, whether linear attention can replicate the ICL capabilities demonstrated by quadratic attention remained an open empirical question that our work directly addresses.

The question of what minimal architectural requirements enable ICL has been explored through increasingly constrained models. Zhang et al. showed in-context learning qualities in the smallest possible transformer with just a single linear self-attention layer trained on linear regression tasks [9]. They proved that if trained correctly, even such a model performs in-context prediction with accuracy competitive with the best linear predictor for the task distribution. However, they showed that this model was brittle to distribution shifts, a weakness that non-linear transformers are better able to overcome. Zhang et al. extended this analysis by examining training dynamics of multi-head linear self-attention models, looking at two different parameterizations of the key and query matrices: a merged parameterization and separate parameterization of both matrices [10]. In the merged parameterization case, training led to a single abrupt transition into an ICL solution, whereas the separate parameterization leads to a more gradual progression. They found that, during training, linear attention models progressively implement principal component regression in-context. These findings suggest that linear attention can achieve ICL, but its training dynamics and learned algorithms may differ fundamentally from quadratic attention.

Beyond architectural mechanics, recent work has examined ICL from complementary theoretical perspectives. Ren et al. approached in-context learning from a representational-learning perspective [5]. They derived a dual model corresponding to a single softmax attention layer -- essentially a separate model whose training dynamics mirror the inference procedure of the attention layer -- and showed that the inference procedure of the attention layer mirrors the training process of the dual model. This connection allowed them to analyze the training dynamics of the dual model and obtain generalization bounds depending on the number of in-context demonstration tokens. They then proposed modifications for deeper transformers inspired by contrastive representation learning to improve ICL capabilities. This dual-model perspective suggests that ICL can be understood through the lens of meta-learning theory, providing a complementary view to the gradient-descent interpretation.

The emergence and stability of ICL during training is not always straightforward. Singh et al. investigated why ICL sometimes appears early in training but later fades [6]. They showed that ICL in transformers competes with context-constrained-in-weights learning (CIWL), which partially solves the task by storing information in the weights rather than in-context. CIWL and ICL can initially support each other but eventually CIWL outcompetes ICL, causing it to disappear. Their analysis provides a mechanistic explanation for this transient behavior, and identifies training conditions under which ICL can remain stable and persistent. Understanding this competition between ICL and CIWL is crucial for interpreting our results, as differences in training dynamics between quadratic and linear attention may affect which learning strategy dominates.

Finally, ICL is not a monolithic capability but rather emerges from multiple distinct mechanisms. Daniels et al. showed that successful in-context learning required both identifying which example in the prompt is relevant and then continuing the pattern associated with that example [2]. Their analysis showed that these two abilities do not emerge simultaneously. Instead, transformers learn to continue a pattern relatively early in training, and the ability to recognize relevant examples develops much later. This separation implies that in-context learning is not a standalone capability, but is the result of multiple underlying mechanisms that evolve at different rates during training. The authors further observed similar results in large language models, showing it generalizes beyond a toy setting. This decomposition of ICL into sub-capabilities suggests that our evaluation should consider not just final performance but also convergence dynamics, as different attention mechanisms may acquire these sub-capabilities at different rates.

\subsection{Our Contribution}

Building on this rich foundation of mechanistic understanding and architectural analysis, our paper empirically compares quadratic and linear attention on the Garg et al. linear regression ICL task. We evaluate both architectures on not just final MSE, but also convergence, emergence of ICL capabilities as a function of model depth and also generalization abilities to different distributions. While prior work has established that both attention types \textit{can} perform ICL, our systematic comparison reveals \textit{how} architectural choices influence the emergence, efficiency, and robustness of in-context learning. Our work sheds light on the extent to which linear attention can replicate the behaviors of quadratic attention and how architectural choices influence the emergence of ICL.

\section{Methods}

\subsection{Task and Data Generation}

We focused on in-context learning of linear function classes. Similar to Garg et al., we define a prompt $P$ as a sequence of input-output pairs followed by a query: 
\[
P = [(x_1, y_1), \dots, (x_k, y_k), x_{query}]
\]
where $x_i \sim \mathcal{N}(0, I_{d_x})$ and $y_i = f(x_i) = w^T x_i$, with $w$ sampled from a distribution $\mathcal{D}$. We generated $w \sim \mathcal{N}(0, I_{d_x})$. Due to computational constraints, we aimed for a smaller model and simplified the task to use input dimension $d_x=5$.

To also evaluate generalization capabilities, we generated an anisotropic dataset where the inputs are drawn from a distribution with a diagonal covariance matrix $\Sigma$:
\[
x_i \sim \mathcal{N}(0, \Sigma), \quad \Sigma = \text{diag}([0.5, 1, 1.5, 1, 1.75])
\]
This effectively scales the variance coordinate-wise. We used $k = 10$, providing 10 context examples of $(x, y)$ pairs before the final $x_{query}$.

\subsection{Model Architectures}

We compare two architectural families: quadratic attention transformers and linear attention transformers, evaluated at three depths (1, 3, and 6 layers).

\paragraph{Shared Hyperparameters}
All models are trained under identical configurations:
\begin{itemize}
    \item Input dimension ($d_x$): 5
    \item Embedding size ($d_{model}$): 256
    \item Attention heads: 4
    \item MLP expansion ratio: 4$\times$ (1024 hidden units)
    \item Optimizer: Adam
    \item Gradient clipping: max-norm 1.0
\end{itemize}

Garg et al. used 12 layers, 8 attention heads, and a 256-dimensional embedding space with 20-dimensional inputs. Since our problem is scaled down ($d_x=5$), we reduced the architecture to 4 heads and examined depths of 1, 3, and 6 layers while maintaining $d_{model}=256$.

We varied the learning rate, batch size, and training steps for each architecture and depth based on our hyperparameter tuning (see Appendix). We selected the best model evaluated by validation loss.

\subsubsection{Quadratic Attention Transformer}

Following Garg et al., we use a decoder-only GPT-2 architecture with 4 attention heads. We rely on causal masking so that the transformer only relies on pairs seen prior to the current position.

Notably, quadratic attention has a runtime complexity of $O(T^2 d_{model})$, where $T$ is the sequence length (number of tokens in the prompt) and $d_{model}$ is the embedding dimension.

\paragraph{Hyperparameters}
We used a learning rate of $1e-4$ and a batch size of $32$. We trained these models for 30,000 steps.

\subsubsection{Linear Attention Transformer}

We implement a custom causal linear attention architecture that replaces the quadratic softmax attention in GPT-style models with an $O(T d_{model}^2)$ kernelized attention mechanism.

\paragraph{Kernelized Attention and Feature Map Selection}
We employ a kernel feature map $\phi(\cdot)$ to decompose the attention matrix. Our architecture treats this mapping as a modular component, allowing batches to evaluate various kernel functions to approximate the softmax geometry.

The choice of $\phi(\cdot)$ is critical for stability. Initial experiments with the identity map $\phi(x) = x$ failed to guarantee non-negativity. We subsequently adopted the squared-ReLU feature map, $\phi(x) = \mathrm{ReLU}(x)^2$. This choice preserves non-negativity and sharpens attention weights, strictly dominating standard ReLU baselines in our preliminary tests.

Let $q_t, k_t, v_t \in \mathbb{R}^{d_{head}}$ represent the query, key, and value vectors for a single head at time step $t$, where $d_{head} = d_{model} / \text{num\_heads}$. To maintain numerical stability, we apply a scaling factor of $d_{head}^{-1/2}$ prior to the feature map:
\[
q'_t = \phi\left(\frac{q_t}{\sqrt{d_{head}}}\right), \qquad k'_t = \phi\left(\frac{k_t}{\sqrt{d_{head}}}\right).
\]

\paragraph{Recurrent Computation}
We compute the causal attention output using a recurrence relation rather than materializing the full $T \times T$ attention matrix (where $T$ is the sequence length). We define two running states: a matrix $S_t \in \mathbb{R}^{d_{head} \times d_{head}}$ aggregating key-value outer products, and a vector $z_t \in \mathbb{R}^{d_{head}}$ for normalization:
\[
S_t = \sum_{j=1}^t k'_j v_j^\top, \qquad z_t = \sum_{j=1}^t k'_j.
\]
The attention output $o_t$ is:
\[
o_t = \frac{q'_t S_t}{q'_t \cdot z_t + \varepsilon},
\]
where $\varepsilon=10^{-6}$ is added for numerical stability.

\paragraph{Hyperparameters}

We trained 3 models with 1, 3, and 6 attention layers. We used a learning rate of $3e-4$ and a batch size of $64$, which we found via hyperparameter tuning.

We varied the number of steps of training based on the number of layers

1 attention layer: 7500 steps\\
3 attention layers: 8000 steps\\
6 attention layers: 10000 steps\\

\subsection{Training and Evaluation}

Our in-context learning objective is to predict $y_k$ given the prompt $P$. During training, we generate predictions for all intermediate points and evaluate the squared error loss. However, for the test set, we strictly evaluate the squared error of the prediction for the final query point $\hat{y}_{query}$.

The final loss function evaluated on the test set is:
\[
\mathcal{L} = \frac{1}{N_{test}} \sum_{i=1}^{N_{test}} \frac{(\hat{y}_{query}^{(i)} - y_{query}^{(i)})^2}{d_x}
\]
where $N_{test}$ is the number of test prompts evaluated, and $d_x=5$ is the input dimension.

We normalize by $d_x$ (as per Garg et al.) to analyze the average error per dimension. This normalization allows for fair comparison, as unnormalized MSE is expected to scale naturally with the dimensionality of the problem.

\section{Hypotheses}

\begin{enumerate}
    \item We expect quadratic attention to perform better than linear attention and have lower MSE. This is because quadratic attention is able to learn more rich features, whereas linear attention is limited by the feature map we use to replace the softmax.
    \item We expect quadratic attention to generalize better to anisotropic distribution shifts for the same reason above.
    \item We expect better performance and generalization as the depth increases for both attention mechanisms. As depth increases, we expect the models to exhibit stronger ICL capabilities as they are able to learn more rich features. Based on the background review, additional layers enable the transformer to perform more steps of gradient descent and achieve a lower loss.
    \item We expect linear attention to have faster convergence than quadratic attention. Linear attention has limited expressivity as compared to quadratic attention, so learning simpler functions should take fewer steps.
    
\end{enumerate}

\section{Results}

We run six total experiments: two architectures (quadratic vs.\ linear attention) each with depths 1, 3 and 6 layers. To ensure robustness of our results, each model is initialized and trained with 5 different seeds. We also report the uncertainty in our losses in both the training curves and the final results. Additionally, we assume the Ordinary Least Squares solution as a baseline, which achieves an MSE of zero in these noise-free linear regression tasks. 

For context, Garg et al. (2022) reported a normalized MSE of approximately 0.02 for their 12-layer transformer trained on 20-dimensional data, which serves as a practical neural baseline for our experiments.

\Needspace{12\baselineskip}
\subsection{Quadratic Attention Transformer}

\begin{table}[H]
\centering
\caption{Quadratic Attention Transformer performance}
\begin{tabular}{lcccc}
\toprule
Config & Params & Train Loss & Test Loss & Anisotropic Test Loss \\
\midrule
1-layer & 13.7 M & 0.7976 ± 0.0543 & 0.8183 ± 0.0688 & 0.8446 ± 0.0656 \\
3-layer & 15.2 M & 0.0502 ± 0.0063 & 0.0894 ± 0.0097 & 0.0936 ± 0.0112 \\
6-layer & 17.6 M & 0.0203 ± 0.0016 & 0.0365 ± 0.0041 & 0.0398 ± 0.0059 \\
\bottomrule
\end{tabular}
\end{table}

\begin{figure}[H]
    \centering
    
    % --- Row 1, Image 1 ---
    \begin{subfigure}[b]{0.49\textwidth}
        \centering
        \lossplot{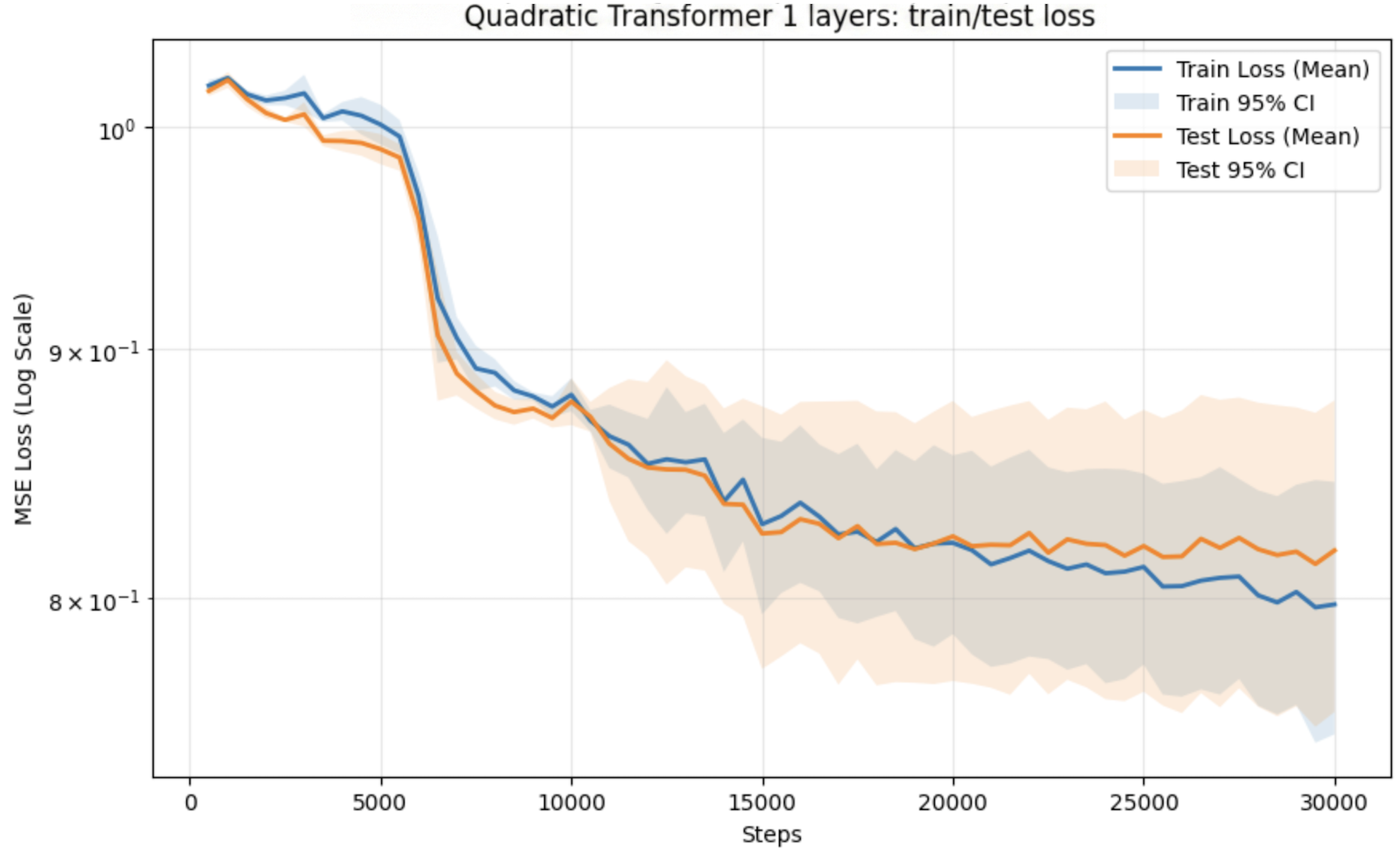}
        \caption{1 Layer Model}
        \label{fig:qt1}
    \end{subfigure}
    \hfill % Pushes Image 1 and 2 to the edges
    % --- Row 1, Image 2 ---
    \begin{subfigure}[b]{0.49\textwidth}
        \centering
        \lossplot{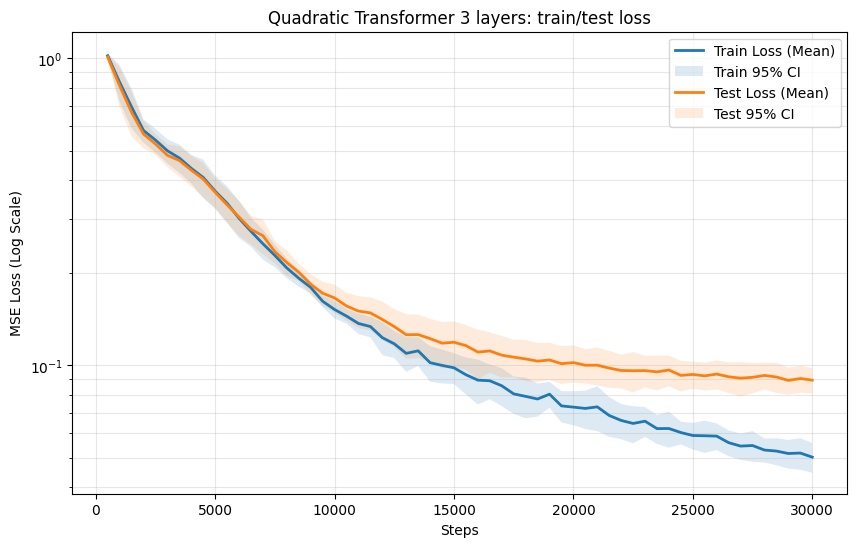}
        \caption{3 Layer Model}
        \label{fig:qt2}
    \end{subfigure}
        
    % --- Row 2, Image 3 (Centered) ---
    \begin{subfigure}[b]{0.5\textwidth}
        \centering
        \lossplot{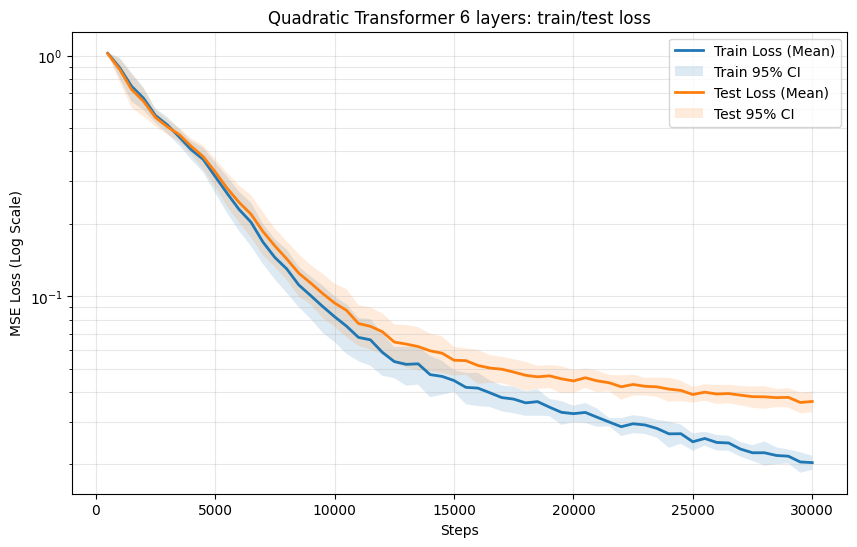}
        \caption{6 Layer Model}
        \label{fig:qt3}
    \end{subfigure}

    \caption{Training and Testing loss. Top row shows shallower networks (1 and 3 layers), while the bottom row shows the deeper 6-layer model.}
    \label{fig:all_qt_results}
\end{figure}

\Needspace{14\baselineskip}
\subsection{Linear Attention Transformer}

\begin{table}[H]
\centering
\caption{Linear Attention Transformer performance}
\begin{tabular}{lcccc}
\toprule
Config & Params & Train Loss & Test Loss & Anisotropic Test Loss \\
\midrule
1-layer & 0.8 M & 0.5946 ± 0.0483 & 0.7430 ± 0.0671  &  0.7867 ± 0.0488 \\
3-layer & 2.38 M & 0.0366 ± 0.0021 & 0.0959 ± 0.0026 & 0.1080 ± 0.0035  \\
6-layer & 4.75 M & 0.0102 ± 0.0012 & 0.0302 ± 0.0034 & 0.0328 ± 0.0030 \\

\bottomrule
\end{tabular}
\end{table}

\begin{figure}[H]
    \centering
    
    % --- Row 1: 1-Layer and 3-Layer ---
    \begin{subfigure}[b]{0.49\textwidth}
        \centering
        \lossplot{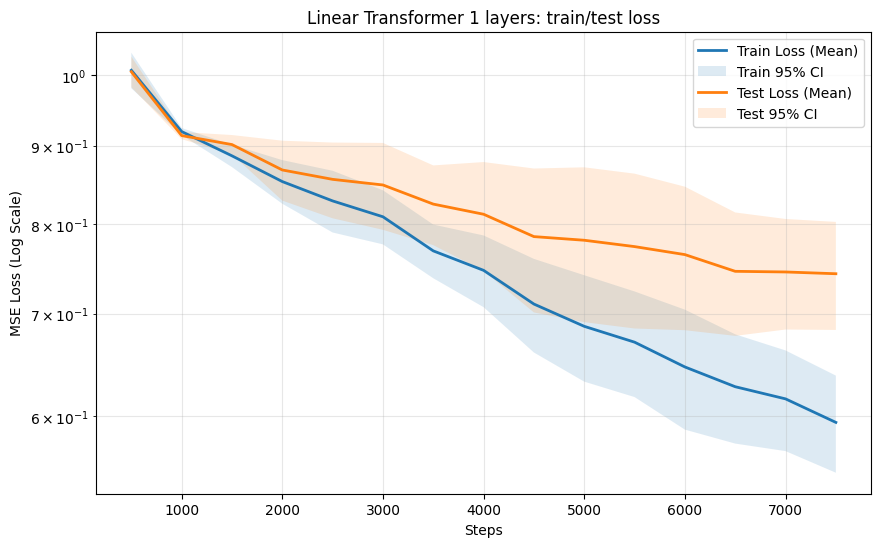}
        \caption{1 Layer Model}
        \label{fig:lt1}
    \end{subfigure}
    \hfill 
    \begin{subfigure}[b]{0.49\textwidth}
        \centering
        \lossplot{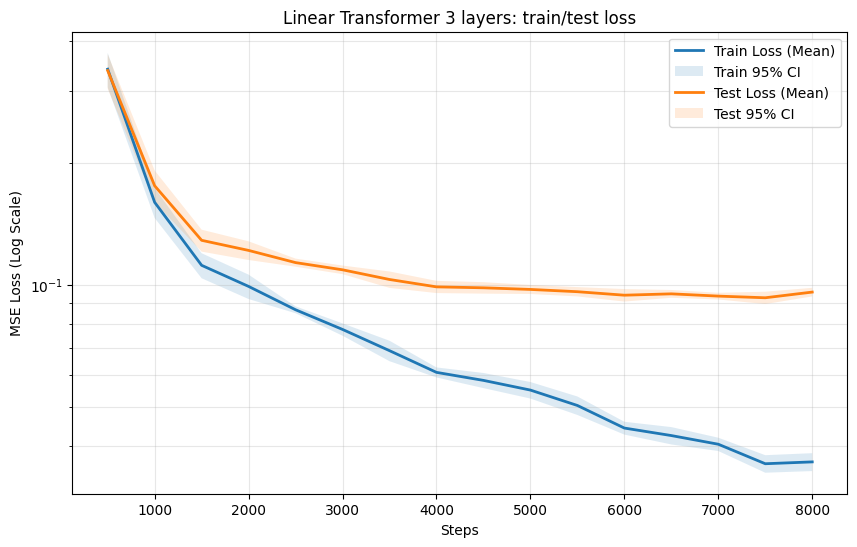}
        \caption{3 Layer Model}
        \label{fig:lt2}
    \end{subfigure}
        
    % --- Row 2: 6-Layer (Centered) ---
    \begin{subfigure}[b]{0.49\textwidth}
        \centering
        \lossplot{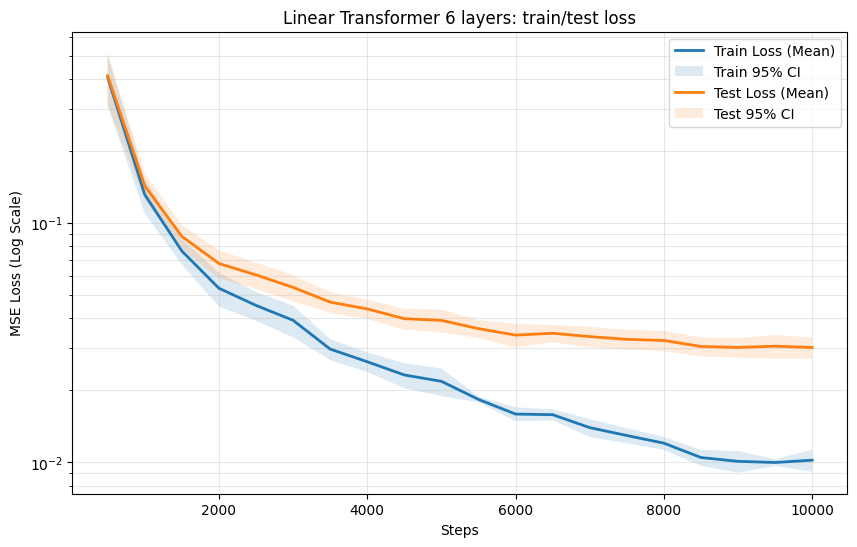}
        \caption{6 Layer Model}
        \label{fig:lt3}
    \end{subfigure}

    \caption{Training and Testing loss for the Linear Attention Transformer}
    \label{fig:all_lt_results}
\end{figure}

\Needspace{18\baselineskip}
\subsection{Robustness to Anisotropic Distributional Shifts}

\begin{figure}[H]
    \centering
    \begin{subfigure}[b]{0.9\linewidth}
        \centering
        \barplot{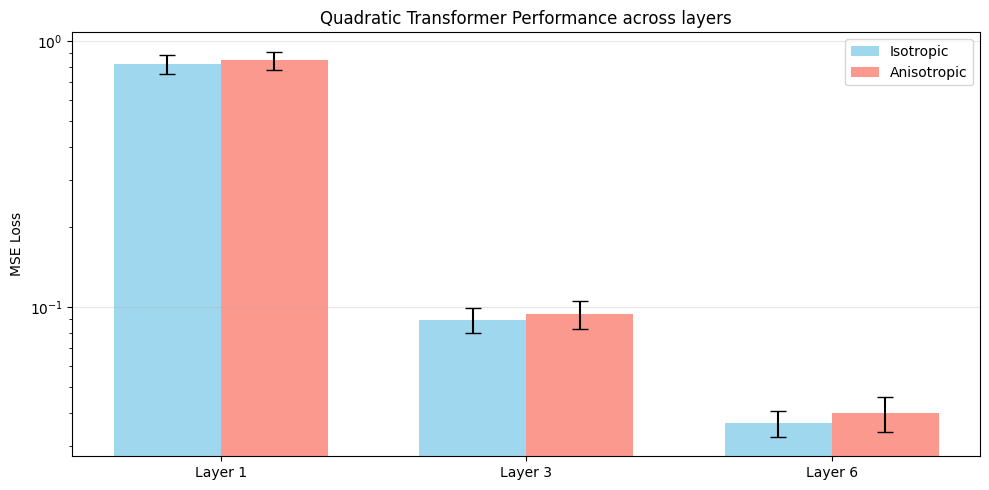}
        \caption{Quadratic Attention performance across Isotropic and Anisotropic Datasets}
        \label{fig:qtcomp}
    \end{subfigure}

    \begin{subfigure}[b]{0.9\linewidth}
        \centering
        \barplot{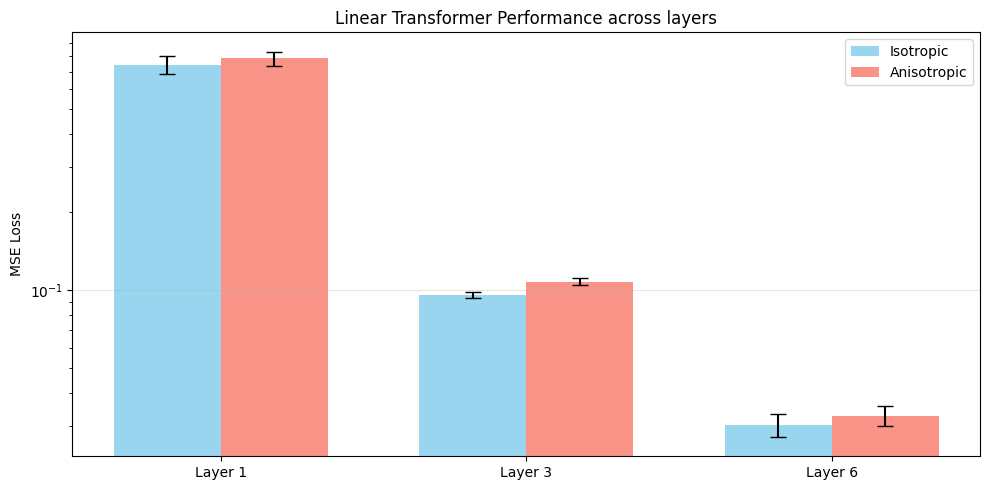}
        \caption{Linear Attention performance across Isotropic and Anisotropic Datasets}
        \label{fig:ltcomp}
    \end{subfigure}

    \caption{Isotropic vs. anisotropic performance across depths.}
    \label{fig:robustness_bars}
\end{figure}

\section{Discussion}

\subsection{Convergence Behavior}

To quantify convergence speed, we measure the number of seen (num steps * batch size) required to reach 90\% of final MSE, defined as reducing 90\% of the gap between initial test loss and final test loss. Table~\ref{tab:conv_steps} shows that linear attention converges substantially faster than quadratic attention across all depths.

\begin{table}[H]
\centering
\caption{Steps and equivalent samples seen to reach 90\% convergence for each architecture and depth.}
\label{tab:conv_steps}
\begin{tabular}{lccc}
\toprule
Model & Batch Size & Steps to 90\% & Samples to 90\% \\
\midrule
Quadratic 1 Layer & 32 & 7000  & 224,000 \\
Quadratic 3 Layer & 32 & 21,500 & 688,000 \\
Quadratic 6 Layer & 32 & 25,000 & 800,000 \\
Linear 1 Layer    & 64 & 4000  & 256,000 \\
Linear 3 Layer    & 64 & 4000  & 256,000 \\
Linear 6 Layer    & 64 & 7500  & 480,000 \\
\bottomrule
\end{tabular}
\end{table}

For the 3-layer and 6-layer models, linear attention reaches 90\% convergence in 256k and 480k samples seen respectively, while quadratic attention requires 688k and 800k seen (Figures~1 and~2). This represents an approximately 1.6$\times$ speedup for linear attention when looking at 6 layers. The training curves confirm this pattern: linear attention models show sharp initial drops in test loss initially before plateauing (Figures~2b and~2c), while quadratic attention models exhibit slower, more gradual convergence that continues throughout the full 30,000 training steps (Figures~1b and~1c). This faster convergence of linear attention likely stems from its more constrained hypothesis and reduced parameter count.

\subsection{Depth Scaling}

Increasing model depth from 1 to 3 to 6 layers dramatically improves ICL capabilities for both architectures, though with different scaling characteristics. For quadratic attention, depth scaling yields progressive improvements: test loss decreases by 89\% as depth increases from 1 to 3 layers and 59\% from 3 to 6 layers (Table~1). 

Linear attention exhibits similar trends. Test loss decreases by 87\% as depth increases from 1 to 3 and by 68\% as depth increases from 3 to 6 (Table~2). However, unlike quadratic attention, the 6-layer linear model requires slightly more training steps to surpass the 3-layer model's performance. The 3-layer model achieves its best performance around 5,000 steps, while the 6-layer model continues improving through 8,000--10,000 steps (Figures~2b and~2c). This suggests that while additional depth increases the model's ultimate capacity, the constrained feature space of linear attention may require more iterations to fully exploit deeper representations.

For the 1-layer models, both architectures struggle significantly and fail to achieve strong in-context learning performance (Tables~1 and~2), with test losses remaining above 0.7 for both. This suggests that a single attention layer, regardless of mechanism, lacks sufficient depth to effectively implement the iterative refinement necessary for robust ICL on linear regression. This aligns with our background research, which shows that one layer is only able to perform a single step of gradient descent, which isn't enough to arrive at the optimal least squares solution.

The most dramatic performance gap occurs at 1 layer, where quadratic attention (test loss 0.8183) slightly outperforms linear attention (0.7430), though both fail to achieve strong ICL. At 3 and 6 layers, both architectures achieve comparable final test performance, with 6-layer linear attention (0.0302) even marginally outperforming 6-layer quadratic attention (0.0365), though the difference falls within uncertainty bounds.

\subsection{Robustness to Distribution Shifts}

Both architectures demonstrate reasonable robustness to the anisotropic distribution shift, though quadratic attention shows slightly better generalization. When evaluated on the anisotropic test set, performance degradation is negligible across all models. For the 6-layer models, quadratic attention's test loss increases from 0.0365 (isotropic) to 0.0398 (anisotropic), representing approximately 9\% degradation, while linear attention increases from 0.0302 to 0.0328, representing approximately 8.6\% degradation (Tables~1 and~2, Figures~3 and~4).

The 3-layer models show similar patterns: quadratic attention degrades from 0.0894 to 0.0936 (4.7\% increase), while linear attention degrades from 0.0959 to 0.1080 (12.6\% increase). This larger gap at 3 layers suggests that quadratic attention may better capture invariances that generalize across input distributions when depth is limited.

Across all depths, the absolute difference between isotropic and anisotropic performance remains small for both architectures, indicating that both learn representations that are reasonably invariant to the scaling of input dimensions. This robustness aligns with findings from Garg et al.\ (2022) that transformers trained on linear regression tasks exhibit strong out-of-distribution generalization. The comparable performance of both architectures on distribution shifts suggests that the choice of attention mechanism (quadratic vs.\ linear) has less impact on generalization than architectural depth and training convergence.

Overall, while quadratic attention maintains a slight edge in distribution robustness, particularly at shallower depths, the differences are relatively minor. Both architectures successfully learn in-context linear regression that generalizes beyond their training distribution, with 6-layer models showing the strongest robustness for both attention mechanisms.

\section{Conclusion}

We were able to successfully replicate the results of Garg et al. on a reduced problem size. We were also able to note the generalization of quadratic attention to anisotropic distribution shifts as seen by them. In addition, we were able to evaluate the performance of linear attention on the same in-context learning task and benchmark its performance and generalization to anisotropic distribution shifts. We found mixed support for our initial hypotheses: while we correctly predicted that linear attention would converge faster (Hypothesis 4), our expectation that quadratic attention would substantially outperform linear attention in final MSE (Hypothesis 1) was not fully supported - at 6 layers, both architectures achieved comparable test performance. 

Our hypothesis regarding depth scaling (Hypothesis 3) was strongly confirmed, with both architectures showing dramatic improvements from 1 to 6 layers. Finally, our prediction about generalization to distribution shifts (Hypothesis 2) was only partially validated: quadratic attention showed marginally better robustness at shallow depths, but the difference diminished at 6 layers.

\section{Future Work}

Several promising directions emerge from our findings. First, we believe that evaluating quadratic and linear attention on more complicated function classes could illustrate differences in learning capabilities and generalization more clearly. Investigating the mechanistic differences between how quadratic and linear attention implement gradient descent or other classical algorithms in-context, building on the work of von Oswald et al. and Akyürek et al., could provide deeper theoretical understanding.

Additionally, our analysis is only limited to two attention mechanisms. We could explore hybrid architectures that combine quadratic and linear attention layers as well. Finally, extending this analysis to state-space models and other efficient attention alternatives would situate linear attention within the broader landscape of scalable sequence modeling architectures. Understanding which architectural properties are essential for ICL versus merely beneficial remains an open and important question for both theoretical understanding and practical deployment of foundation models.

\paragraph{Reproducibility}
All code required to reproduce our experiments is available at:
\url{https://github.com/Yushgoel/icl-benchmarking}
The repository includes:
\begin{itemize}
\item Training scripts for all six model configurations (quadratic vs.\ linear attention, 1, 3 and 6 layers)
\item Data generation utilities for both isotropic and anisotropic linear regression tasks
\item Checkpointing and logging utilities (JSON logs for losses and configurations)
\item Scripts to reproduce all evaluation plots and tables in this paper
\end{itemize}
To ensure experiment reproducibility, all runs use 5 random seeds (42, 100, 7, 10, 2025) for NumPy and PyTorch. We provide configuration files specifying model widths, depths, optimizer settings, and training hyperparameters. All experiments were performed on a single NVIDIA A100 GPU. The repository includes instructions for reproducing training, evaluation, and distribution-shift experiments with a single command.

\clearpage
\section*{References}

\small{
[1] Akyürek, E., Schuurmans, D., Andreas, J., Ma, T., \& Zhou, D. (2022). What learning algorithm is in-context learning? Investigations with linear models. {\it arXiv preprint arXiv:2211.15661}.

[2] Daniels, S., Davis, D., Gautam, D., Liao, W., Ranade, G., \& Sahai, A. (2025). Decomposing Prediction Mechanisms for In-Context Recall. {\it arXiv preprint arXiv:2507.01414}.

[3] Garg, S., Tsipras, D., Liang, P., \& Valiant, G. (2022). What Can Transformers Learn In-Context? {\it OpenReview}. 

[4] Katharopoulos, Angelos. Vyas, Apoorv. Pappas, Nikolaos. \& Fleuret, Francois. (2020). Transformers are RNNs: Fast Autoregressive Transformers with Linear Attention {\it arXiv preprint arXiv:2006.16236}

[5] Ren, R \& Liu, Y. (2024). Towards Understanding How Transformers Learn In-context Through a Representation Learning Lens {\it NIPS 2024}

[6] Singh, A. K., Moskovitz, T., Dragutinović, S., Hill, F., Chan, S. C. Y., \& Saxe, A. (2025). Strategy Coopetition Explains the Emergence and Transience of In-Context Learning. {\it arXiv preprint arXiv:2503.05631}.

[7] von Oswald, J., et al. (2022). Transformers Learn In-Context by Gradient Descent. {\it arXiv preprint arXiv:2212.07677}.

[8] Zhang, R., Frei, S., \& Bartlett, P. L. (2023). Trained Transformers Learn Linear Models In-Context. {\it arXiv preprint arXiv:2306.09927}.

[9] Zhang, Y., Singh, A. K., Latham, P. E., \& Saxe, A. (2025). Training Dynamics of In-Context Learning in Linear Attention. {\it arXiv preprint arXiv:2501.16265}.
}
\newcommand{\mycomment}[1]{}

\mycomment{
\section*{AI Disclosure}
We used generative AI to help modify some of our training and inference code, as well as helping with plotting and keeping track of results across experiments. We also used it to help edit our report.
}

\clearpage

\section*{Appendix}

\subsection*{Hyperparameter Tuning}

There were 3 main hyperparameters we tuned: learning rate, batch size and number of steps.

\textbf{Learning Rate}\\
For learning rate, we tried a grid search method with values ranging from $5e^{-5}$ to $1e^{-3}$. 

For the quadratic attention models, we found that even for learning rate values of $3e^{-4}$ and above, the models would struggle to learn as the updates were too large. For learning rate values smaller than $1e^{-4}$, we found the model would take too long for convergence (over $50,000$ steps). Therefore, we picked $1e^{-4}$ as our final learning rate. We observed these trends for models of all depths, and the same learning rate was the best for the 1, 3 and 6 layer models.

For the linear attention models, we found that a larger learning rate worked better. For values smaller than $3e^{-4}$, the model took longer to converge but didn't show a significant improvement in final MSE. For values larger than $3e^{-4}$, the model started struggling to learn again, so we settled on $3e^{-4}$ as the final learning rate. 

\textbf{Batch Size}\\
We also tuned the batch size, trying values of $16, 32, 64, 128$. We found for linear attention, a batch size of $64$ was optimal, and for quadratic attention, a batch size of $32$ was optimal.

\textbf{Steps}\\
For the final number of steps used, we first trained the models for a large number of steps ($50,000$ steps each) and then used early stopping to stop and pick the best model when it had converged. For quadratic attention we saw that final convergence happened at $30,000$ steps, whereas for linear attention, the number of steps required for convergence varied by model depth.

\textbf{Linear Attention Feature Map}\\
For linear attention, we additionally experimented with various features maps. As expected, we found that the linear feature map $\phi(x) = x$, performed extremely poorly and didn't learn at all (with a final mse near $1$ for even a 6 layer model). We also tried a relu feature map $\phi(x) = relu(x)$ and a quadratic feature map $\phi(x) = \left[1, x_1, x_2, \ldots, x_d, x_1^2, x_1 x_2, \ldots, x_1 x_d, x_2^2, x_2 x_3, \ldots, x_d^2\right]^\top$ to enforce non-negativity like softmax attention. We saw an improvement in MSE with both of these feature maps, and we finally experimented with combining them to use a squared relu feature map, which had the best MSE.

\vspace{1\baselineskip}
\subsection*{Confidence Intervals}
To calculate confidence intervals, we first computed the residuals of the model across all 5 seeds $\hat{e}$. We then concatenated these residuals, and used $1.96 \cdot std(\hat{e})$ to get the 95\% confidence interval.

\mycomment{
\subsection*{Reviewer Comments}

We compare the performance of quadratic attention and linear attention models on the same dataset to benchmark them. We had initially kept the hyperparameters fixed across both architectures for comparison. Our initial reviewers pointed this out to be a problem, as it is possible that one of the architectures could perform significantly better if only it was tuned. We agree with this reasoning, and so tuned each model independently to allow for comparing the "best" models' performance. Our final reviewers saw the different hyperparameters to be a problem as it could be considered changing more than one variable at a time. However, given the hyperparameter tuning process we used, we feel that the different hyperparameters are not confounding factors as if increasing or decreasing the learning rate affected performance, it would've been reflected in the final hyperparameters we used as well. Therefore, we choose to report results for where the models have been independently tuned as we feel this provides a more accurate comparison.

Our initial reviewers also left feedback mentioning the lack of error bars and uncertainty in our MSE, which is especially important when comparing different models and analyzing performance on distribution shifts. We added $95\%$ confidence intervals to show which results are statistically significant and which ones aren't.

Our final reviewers also gave feedback that our literature review didn't flow as a coherent narrative, and didn't always draw relevant connections to our work. We changed our review to provide more context on the work done in in-context learning in the past and how it relates to our paper.

In the poster session, we also received a lot of questions about our hyperparameter tuning process. We realized we hadn't described this in detail in our paper either and could be an area of interest to the reader, so we added a section in the appendix explaining our process and observations.

Also, from reviewers unfamiliar with in-context learning, they also requested benchmarks from previous work as a helpful baseline to keep in mind. We added in the results from the Garg et al paper, to also provide a sanity check that our quadratic attention models were able to show in-context learning capabilities in roughly the same ballpark as them.
}
\end{document}